%% file: main.tex
\documentclass[10pt, conference, compsocconf]{IEEEtran}
\IEEEoverridecommandlockouts

\usepackage{hyperref}
\usepackage[ruled,norelsize]{algorithm2e}

\usepackage{multirow}
\usepackage{amsmath,amssymb,amsfonts}
\usepackage{amsthm}
\usepackage{microtype}
\usepackage{graphicx}
\usepackage{setspace}
\usepackage{multicol}
\usepackage{multirow}
\usepackage{float}
\usepackage{color}
\usepackage{alltt, listings}
\usepackage{subfigure}
\usepackage{algpseudocode}
\usepackage{amssymb}
\usepackage{booktabs}
\usepackage{bm}
\usepackage{balance}
\usepackage{wrapfig}
\usepackage{url}
\usepackage{graphicx}
\usepackage{xcolor}
\usepackage{footnote}
\usepackage[bottom]{footmisc}
\makeatletter
\newcommand{\removelatexerror}{\let\@latex@error\@gobble}
\makeatother
\newcommand{\tabincell}[2]{\begin{tabular}{@{}#1@{}}#2\end{tabular}}  
\newtheorem{theorem}{Theorem}
\newtheorem{lemma}{Lemma}

\newtheorem{assumption}{Assumption}

\def\BibTeX{{\rm B\kern-.05em{\sc i\kern-.025em b}\kern-.08em
    T\kern-.1667em\lower.7ex\hbox{E}\kern-.125emX}}
\begin{document}

\title{DQ-SGD: Dynamic Quantization in SGD for Communication-Efficient Distributed Learning}



\author{\IEEEauthorblockN{Guangfeng Yan\IEEEauthorrefmark{1},
Shao-Lun Huang\IEEEauthorrefmark{2}, Tian Lan\IEEEauthorrefmark{3} and
Linqi Song\IEEEauthorrefmark{1}}
\IEEEauthorblockA{
\IEEEauthorrefmark{1}Department of Computer Science, City University of Hong Kong, Hong Kong SAR\\
\IEEEauthorrefmark{1}City University of Hong Kong Shenzhen Research Institute, Shenzhen, China\\
\IEEEauthorrefmark{2}Data Science and Information Technology Research Center, Tsinghua-Berkeley Shenzhen Institute, Shenzhen, China\\
\IEEEauthorrefmark{3}Department of Electrical and Computer Engineering,
George Washington University, Washington, DC, USA\\
Email: gfyan2-c@my.cityu.edu.hk,
twn2gold@gmail.com,
tlan@gwu.edu,
linqi.song@cityu.edu.hk}}
\maketitle

\input{1.abstract}
\input{2.keyword}
\input{3.introduction}
\input{4.related_work}
\input{5.problem_definition}
\input{6.dqsgd}
\input{6.5.discussion}
\input{7.experiments}
\input{8.conclusion}
\input{Acknowledgment}
\input{9.bibliography}
\input{A.appendix}

\end{document}

%% file: 1.abstract.tex
\begin{abstract}
Gradient quantization is an emerging technique in reducing communication costs in distributed learning. Existing gradient quantization algorithms often rely on engineering heuristics or empirical observations, lacking a systematic approach to dynamically quantize gradients. This paper addresses this issue by proposing a novel dynamically quantized SGD (DQ-SGD) framework, enabling us to dynamically adjust the quantization scheme for each gradient descent step by exploring the trade-off between communication cost and convergence error. We derive an upper bound, tight in some cases, of the convergence error for a restricted family of quantization schemes and loss functions. We design our DQ-SGD algorithm via minimizing the communication cost under the convergence error constraints. Finally, through extensive experiments on large-scale natural language processing and computer vision tasks on AG-News, CIFAR-10, and CIFAR-100 datasets, we demonstrate that our quantization scheme achieves better tradeoffs between the communication cost and learning performance than other state-of-the-art gradient quantization methods.
\end{abstract}

%% file: 2.keyword.tex

%% file: 3.introduction.tex
\section{Introduction}
\label{sec:introduction}

Distributed Stochastic Gradient Descent (SGD) is the core in a vast majority of distributed learning algorithms.
Due to the limited bandwidth in practical networks, communication overhead for transferring gradients often becomes the performance bottleneck. 
Gradient quantization is an effective approach towards communication-efficient distributed learning, which uses fewer number of bits to approximate the original real value  \cite{b4,b7,b8,b9,b10}. The lossy quantization inevitably brings in gradient noise, which hurts the convergence of the model. Hence, a key question is how to effectively select the number of quantization bits to balance the trade-off between the communication cost and the convergence performance. 

Existing algorithms often quantize parameters into a fixed number of bits for all training iterations, which is inefficient in balancing the communication-convergence trade-off. To further reduce the communication overhead, some empirical studies began to dynamically adjust the number of quantization bits according to current model parameters in the training process, such as the gradient’s mean to standard deviation ratio \cite{b11}, the training loss \cite{b12}, gradient’s root-mean-squared value \cite{b13}. Though these empirical heuristics of adaptive quantization methods show good performance in specific tasks, their imprecise conjectures and the lack of theoretical guidelines in the conjecture framework have limited their generalization to a broad range of machine learning models/tasks.

This paper proposes a novel dynamically quantized SGD (DQ-SGD) framework for minimizing communication overhead in distributed learning while maintaining the desired model performance. Under the assumption of smoothness and strong convexity, we first derive an upper bound on the gap between the loss after $T$ iterations and the optimal loss to characterize the convergence error caused by limited iteration steps, sampling, and quantization.
Based on the above theoretical analysis, we design a dynamic quantization algorithm by minimizing the total communication cost under desired model performance constraints. Our dynamic quantization algorithm can adjust the number of quantization bits adaptively by taking into account the desired model performance, the remaining number of iterations, and the norm of gradients. We validate our theoretical analysis through extensive experiments on large-scale Natural Language Processing (NLP) and Computer Vision (CV) tasks, including text classification tasks on AG-News and image classification tasks on CIFAR-10 and CIFAR-100. Numerical results show that our proposed DQ-SGD significantly outperforms the baseline quantization methods.

To summarize, our key contributions are as follows:

$\bullet$ We propose a novel framework to characterize the trade-off between communication cost and convergence error by dynamically quantizing gradients in the distributed learning.

$\bullet$ We derive an upper bound on the convergence error for smooth strongly-convex objectives, and the upper bound is shown to be tight for a special case of quadratic functions with isotropic Hessian matrix.

$\bullet$ We develop a dynamically quantized SGD strategy, which is shown to achieve a fewer communication cost than fixed-bit quantization methods.

$\bullet$ We validate the proposed DQ-SGD on a variety of real-world datasets and machine learning models, demonstrating that our proposed DQ-SGD significantly outperforms state-of-the-art gradient quantization methods in terms of mitigating communication costs.

%% file: 4.related_work.tex
\section{Related Work}
\label{key}
To mitigate the communication bottleneck in distributed SGD, gradient quantization has been investigated. Different fixed number of bits quantization methods have been studied, such as 1BitSGD \cite{b7,b8}, TernGrad (ternary levels) \cite{b14}, QSGD (arbitrary fixed number of bits) \cite{b4}. 

However, these fixed-bit quantization methods may not be efficient in communication; and more efficient schemes that can dynamically adjust the number of quantization bits in different gradient descent step may have the potential to improve the communication-convergence tradeoff performance. Several studies try to construct adaptive quantization schemes through engineering heuristics or empirical observations. However, they do not come up with a solid theoretical analysis \cite{b11,b12,b13}, which even results in contradicted conclusions. More specifically,  MQGrad \cite{b12}, and AdaQS \cite{b13} suggest using few quantization bits in early epochs and gradually increase the number of bits in later epochs; while the scheme proposed by Anders \cite{b11} states that more quantization bits should be used for the gradient with a larger root-mean-squared (RMS) value, choosing to use more bits in the early training stage and fewer bits in the later stage. 

This paper's key contribution is to develop a systematic framework to crystallize the design trade-off in dynamic gradient quantization and settle this contradiction.

%% file: 5.problem_definition.tex
\section{Problem Formulation}
\label{sec:problem_definition}

We consider a distributed learning system with $W$ workers and a parameter server. Data are distributed over $W$ workers, with a shared model to be jointly optimized. The local dataset of worker $i$ is $D_i$. 
We aim to minimize the objective function $F: \mathbb{R}^d \to \mathbb{R} $ with model parameter $\mathbf{x}$
\begin{equation}
    \setlength{\abovedisplayskip}{3pt}
    \setlength{\belowdisplayskip}{3pt}
	\min_{\mathbf{x} \in \mathbb{R}^d}F(\mathbf{x}) = \cfrac{1}{W} \sum_{i=1}^W  \mathbb{E}_{\xi\sim D_i}[l(\mathbf{x};\xi)],
	\label{optim_problem}
\end{equation}
where $l(\mathbf{x};\xi)$ is the loss of the model $\mathbf{x}$ at data point $\xi$. A standard approach to solve this problem is distributed SGD, where each worker $i$ computes its local stochastic gradient at iteration $t$ with model parameter $\mathbf{x}_t$: $\mathbf{g}^{(i)}_t = \nabla l(\mathbf{x}_t;\xi^{(i)})$. Then these local gradients are sent to the parameter server, and the server aggregates these gradients to update the model: $\mathbf{x}_{t+1} = \mathbf{x}_t - \frac{\eta}{W}\sum_{i=1}^W \mathbf{g}^{(i)}_t$, where $\eta$ is the learning rate.
To reduce the communication cost, we consider to quantize the local stochastic gradients before sending them to the server:
\begin{equation}\
	\mathbf{x}_{t+1} = \mathbf{x}_t - \cfrac{\eta}{W}\sum_{i=1}^W \mathcal {Q}_{b_t}[\mathbf{g}^{(i)}_t],
	\label{equ:qsgd}
\end{equation}
where $\mathcal{Q}_{b_t}[\cdot]$ is the quantization operator and $b_t$ is the number of quantization bits at iteration $t$ (in other words, we may allocate a different number of quantization bits at different iteration steps). 

It is clear that the lossy compression inevitably affects the convergence of model training and deteriorates the learning performance. Therefore, we 
use the gap between the loss after $T$ iterations and the optimal loss to characterize the learning performance. We say the algorithm achieves an $\epsilon \text{-suboptimal solution}$ if
\begin{equation}
	F(\mathbf{x}_T) - F(\mathbf{x}^*) \le \epsilon
\end{equation}
where $\mathbf{x}^*$ is the optimal solution to minimize $F$. Note that this suboptimal gap $\epsilon$ depends not only on the constrained communications between workers and servers, but also on the limited number of iterations $T$, the stochastic sampling, and the initial model parameter.

In this work, given the total number of training iterations $T$, the number of workers $W$, and the desired model performance $\epsilon$, we aim to adaptively adjust the number of quantization bits $b_t$ for each step to minimize the total communication cost under the model performance constraints.

Formally, {\bf our design of DQ-SGD is to solve the following \emph {Dynamic Quantization Problem (DQP)}:}
\begin{equation}
\begin{split}
 	\text{(DQP):}\quad & \min_{\{b_t\}} f_\mathcal{Q}(T, W, \{b_t\})\\
	& s.t. \quad F(\mathbf{x}_T) - F(\mathbf{x}^*) \le \epsilon,
	\label{eq:DQP}   
\end{split}
\end{equation}
where $f_\mathcal{Q}(T, W, \{b_t\})$ is the incurred total communication cost of $T$ iterations and our goal is to find appropriate dynamic quantization schemes $\{b_t\}$ for $T$ iterations.


%% file: 6.dqsgd.tex
\section{Dynamically Quantized SGD}
\label{sec:dqsgd}

In general, the DQP problem is not easy to solve and relaxations are needed to approach this problem. Therefore, we propose to solve a relaxed version of the DQP problem and design a DQ-SGD algorithm based on the solution, which we show performs sufficiently well in practice in the experiments.

More specifically, we relax the DQP problem from the following two perspectives.

$\bullet$ We restrict our quantization scheme to a family of Element-Wise Uniform (EWU) quantization schemes, which are unbias with bounded variance.

$\bullet$ We relax the constraint in Eq.~\eqref{equ:qsgd} to upper bound the convergence rate for smooth strongly-convex loss functions.

\subsection{Element-Wise Uniform Quantization}

There are several types of quantization operations -- categorized from different perspectives, such as grid quantization, uniform and non-uniform quantization, biased and unbiased quantization. Here, we adopt a family of stochastic quantization --EWU, similar to \cite{b4}, to quantize the gradients. 

In this EWU scheme, The $j$-th component of the stochastic gradient vector $\mathbf{g}$ (for any worker $i$) is quantized as
\begin{equation}\
	\mathcal{Q}_b[g_j] = \|\mathbf{g}\|_p \cdot \text{sgn}(g_j)\cdot\zeta(g_j,s),
	\label{quantizedoperation}
\end{equation} 
where $\|\mathbf{g}\|_p$ is the $l_p$ norm of $\mathbf{g}$; $\text{sgn}(g_j)=\{+1,-1\}$ is the sign of $g_j$; $s$ is the quantization level. Note that, the quantization level is roughly exponential to the number of quantized bits. If we use $b$ bits to quantize $g_j$, we will use one bit to represent its sign and the other $b-1$ bits to represent $\zeta(g_j,s)$, thus resulting in a quantization level $s=2^{b-1}-1$.  And $\zeta(g_j,s)$ is an unbiased stochastic function that maps scalar ${\vert g_j \vert}/{\|\mathbf{g}\|_p}$ to one of the values in set $\{0, 1/s, 2/s, \ldots, s/s\}$: if ${\vert g_j \vert} / {\|\mathbf{g}\|_p} \in [l/s,(l+1)/s]$, we have
\begin{equation}\
	\zeta(g_j,s) = \begin{cases} l/s,& \text{with probability $1-p_r$,}\\ 
	(l+1)/s,& \text{with probability $p_r = s\cfrac{\vert g_j \vert}{\|\mathbf{g}\|_p}-l$.} \end{cases}
\label{eq:map}
\end{equation} 
Hence, the incurred total communication cost is:
\begin{equation}
	f_{\mathcal{Q}}(T, W, \{b_t\}) = W\sum_{t=0}^{T-1} \left[db_t+B_{pre}\right],
\end{equation}
where $B_{pre}$ is the number of bits of full-precision floating point (e.g., $B_{pre}=32$ or $B_{pre}=64$) to represent $\|\mathbf{g}\|_p$. If we make the commonly used assumption for stochastic gradients as follow:

\begin{assumption}[Unbiasness and Bounded Variance of Stochastic Gradient]
	The stochastic gradient oracle gives us an independent unbiased estimate $\mathbf{g}$ with a bounded variance:
	\begin{equation}
		\mathbb{E}_{\xi\sim D_i}[\mathbf{g}^{(i)}_t] = \nabla F(\mathbf{x}_t),
	\end{equation}
	\begin{equation}
		\mathbb{E}_{\xi\sim D_i}[\|\mathbf{g}^{(i)}_t-\nabla F(\mathbf{x}_t)\|_2^2] \le \sigma^2.
	\end{equation}
	\label{ass:stochastic_gradient} 
\end{assumption}
Then we have the following lemma to characterize the aggregated stochastic gradient ${\mathbf{\hat g}}_t \triangleq \cfrac{1}{W}\sum_{i=1}^W \mathcal {Q}_{b_t}[\mathbf{g}^{(i)}_t]$, and the proof is given in Appendix \ref{pro: lem1}. 

\begin{lemma}[Unbiasness and Bounded Variance of EWU]
	For the local gradient $\mathbf{g}^{(i)}_t$, if the number of quantization bits for all $W$ workers are all $b_t$, then the aggregated gradient ${\mathbf{\hat g}}_t$ satisfies: 
	\begin{equation}
		\mathbb{E}[{\mathbf{\hat g}}_t] = \nabla F(\mathbf{x}_t)
		\label{eq:unbiassness}
	\end{equation}
	and
	\begin{equation}
	\begin{split}
	  \mathbb{E}\left[||\mathbf{\hat g}_t||_2^2\right] \le \|\nabla F(\mathbf{x}_t)\|_2^2 
	   + \underbrace{\cfrac{\sigma^2}{W}}_{\text{\rm Sampling Noise}} + \underbrace{\cfrac{d}{4W(2^{b_t-1}-1)^2}\bar G_t^2}_{\text{\rm Quantization Noise}},
	\end{split}
	\label{eq:qsg}
	\end{equation}
	where $\bar G_t^2 = \cfrac{1}{W}\sum_{i=1}^W\|\mathbf{g}^{(i)}_t\|_p^2$, is the mean square of all local gradient $l_p$ norms at iteration $t$.
	\label{lem:qsg}
\end{lemma}

Eq.~\eqref{eq:unbiassness} means that the aggregated gradient ${\mathbf{\hat g}}_t $ is the unbiased estimate of $\nabla F(\mathbf{x})$. Eq.~\eqref{eq:qsg} implies that the difference between $||\mathbf{\hat g}_t||_2^2$ and $\|\nabla F(\mathbf{x}_t)\|_2^2$ consists of two parts: the first part is the sampling noise, which inversely proportional to $W$; the second part is the quantization noise, which 
is proportional to $\bar G_t^2$ and decays exponentially with the increase of the number quantization bits $b_t$.


\subsection{Upper Bounded Convergence Rate for Smooth and Strongly Convex Functions}
We first state some assumptions as follows.
\begin{assumption}[Smoothness]
	The objective function $F(\mathbf{x})$ is $L$-smooth, if $\forall \mathbf{x},\mathbf{y} \in \mathbb{R}^d$, $\|\nabla F(\mathbf{x})-\nabla F(\mathbf{y})\|_2 \leq L\|\mathbf{x}-\mathbf{y}\|_2$.
	\label{ass:smoothnesee} 
\end{assumption}
It implies that $\forall \mathbf{x},\mathbf{y} \in \mathbb{R}^d$, we have
\vspace{-0.1in}
\begin{equation}
	F(\mathbf{y}) \leq F(\mathbf{x}) + \nabla F(\mathbf{x})^\mathrm{T} (\mathbf{y}-\mathbf{x}) + \cfrac{L}{2} \|\mathbf{y}-\mathbf{x}\|_2^2
\end{equation}
\begin{equation}
	\|\nabla F(\mathbf{x})\|_2^2 \leq 2L [F(\mathbf{x}) - F(\mathbf{x}^*)]
	\label{eq:smoothnesee_norm}
\end{equation}

\begin{assumption}[Strong convexity]
	The objective function $F(\mathbf{x})$ is $\mu$-strongly convex, if $\exists \mu > 0$, $F(\mathbf{x}) - \cfrac{\mu}{2} \mathbf{x}^\text{\rm T}\mathbf{x}$ is a convex function.
	\label{ass:strongly convexity} 
\end{assumption}

From Assumption \ref{ass:strongly convexity}, we have: $\forall \mathbf{x} \in \mathbb{R}^d$,
\begin{equation}
	\|\nabla F(\mathbf{x})\|_2^2 \geq 2\mu [F(\mathbf{x}) - F(\mathbf{x}^*)]
	\label{eq:Strongconvex_norm}
\end{equation}

Putting the quantized SGD (\ref{equ:qsgd}) on smooth, strongly convex functions yield the following result with proof given in Appendix \ref{pro:Theorem 1}.

\begin{theorem}[Convergence Error Bound of Strongly Convex Objectives]
	For the problem in Eq. \eqref{optim_problem} under Assumption \ref{ass:smoothnesee}, \ref{ass:strongly convexity} and \ref{ass:stochastic_gradient} with initial parameter $\mathbf{x}_0$, using quantized gradients in Eq.~\eqref{equ:qsgd} for iteration, we can upper bound the convergence error by
\begin{equation}
\begin{split}
	&\mathbb{E}[F(\mathbf{x}_T)-F(\mathbf{x}^*)]\\
    &\le \underbrace{\alpha(\eta)^T [F(\mathbf{x}_0)-F(\mathbf{x}^*)] + \cfrac{L\eta^2\sigma^2[1-\alpha(\eta)^T]}{2W(1-\alpha(\eta))}}_{\text{\rm Error of Distributed SGD}}\\  
    &\quad + \underbrace{\cfrac{Ld\eta^2}{8W} \sum_{t=0}^{T-1} \alpha(\eta)^{T-1-t}\cfrac{\bar G_t^2}{(2^{b_t-1}-1)^2}}_{\text{\rm Quantization Error}}\\
    &\overset{T \to \infty}{\rightarrow} \cfrac{L\eta^2\sigma^2}{2W(1-\alpha(\eta))} + \cfrac{Ld\eta^2}{8W} \sum_{t=0}^{T-1} \alpha(\eta)^{T-1-t}\cfrac{\bar G_t^2}{(2^{b_t-1}-1)^2}
    \label{eq:Theorem 1}
    \end{split}
    \end{equation}
	where $\alpha(\eta) =: 1-2\mu\eta+L\mu\eta^2$ (We abbreviate $\alpha(\eta)$ as $\alpha$ in the following section.). \label{Theorem 1}
\end{theorem}

We can see that the convergence error consists of two parts: the first two terms are the error of the distributed SGD method, which is independent of the quantization algorithms. This part error can be reduced by increasing the number of iterations $T$ and also depends on the learning rate $\eta$ (from the expression of $\alpha$, we can see that when $\eta \le 1/L$, with the increase of $\eta$, $\alpha$ decrease, and the convergence rate of the model is accelerated); The last term is \textbf{quantization error}, resulted from the lossy quantization of gradients. It is obtained by the weighted accumulation of quantization noise at each iteration and directly increases the convergence error floor. Note that $\alpha$ is less than 1. Thus the weight given to quantization noise decays exponentially as the number of intervening iterations increases. Accordingly, this is sometimes called an \textbf{exponential recency-weighted average}. 

We can see that the quantization error decays exponentially in the number of quantization bits $b_t$. When the number of quantization bits at each iteration is large enough (e.g., $b_t = 32$), the quantization error tends to 0, but the communication cost is significantly high. 

We aim to use as little communication cost as possible to ensure that the quantization error is below a given level $\epsilon_{\mathcal{Q}} = [1-\gamma]\epsilon$, where $1-\gamma$ is a tradeoff factor representing the contribution of convergence error by quantization. Furthermore, we have $\lim_{T \to \infty} \epsilon_{\mathcal{Q}} = \epsilon - \frac{L\eta^2\sigma^2}{2W(1-\alpha)}$. 

\subsection{DQ-SGD Algorithm}

Given the above two relaxations, we can rewrite the DQP as
\begin{equation}
\begin{split}
    &\min_{\{b_t\}}\quad W\sum_{t=0}^{T-1} (db_t+B_{pre}),\\
	& s.t. \quad  \cfrac{Ld\eta^2}{8W} \sum_{t=0}^{T-1} \alpha^{T-1-t} \cfrac{\bar G_t^2}{(2^{b_t-1}-1)^2} = \epsilon_{\mathcal{Q}}.
\label{eq:optimization}
\end{split}
\end{equation}
By solving the above optimization problem, we can determine the $\{b_t\}$ at every iteration step:
\begin{equation}
	b_t = \log_2{\left[\sqrt{\cfrac{T}{{\hat \epsilon}_{\mathcal{Q}}}}\alpha^{(T-1-t)/2}\bar G_t + 1\right]}+1
	\label{eq:bits}
\end{equation}
where ${\hat \epsilon}_{\mathcal{Q}} \triangleq \cfrac{8W}{Ld\eta^2}\epsilon_{\mathcal{Q}}$. We can see that the number of quantization bits is determined by three key factors: (i) the desired quantization error upper bound $\epsilon_{\mathcal{Q}}$, the smaller the desired quantization error is, the more quantization bits are needed; (ii) the iteration step $t$, the number of bits is increasing as the training process goes on; (iii) the root-mean-square of local gradient norms $\bar G_t$, gradients with a larger norm should be quantized using more bits. 

The pseudocode is given in Algorithm \ref{alg:DQSGD in distributed learning}. We have a set of $W$ workers who proceed in synchronous steps, and each worker has a complete copy of the model. In each communication round, workers compute their local gradients and communicate quantized gradients with the parameter server (lines 3-5), while the server aggregates these gradients from workers and updates the model parameters (lines 8-10). If $\mathcal {Q}_{b_t}[\mathbf{g}^{(i)}_t]$ is the quantized stochastic gradients in the $i$-th worker and $\mathbf{x}_t$ is the model parameter that the workers hold in iteration $t$, then the updated value of $\mathbf{x}$ by the end of this iteration is: $\mathbf{x}_{t+1} = \mathbf{x}_t + \eta \mathbf{\hat g}_t$, where $\mathbf{\hat g}_t = \frac{1}{W}\sum_{l=1}^W \mathcal {Q}_{b_t}[\mathbf{g}^{(i)}_t]$. Note that we determine $b_{t+1}$ according to the gradients information at iteration $t$ (lines 11), so we update the compression bits every $
\tau$ iterations in practice (In our experience, we take $\tau = 100$).

\textbf{DQ-SGD outperforms fixed-bit quantization based SGD in communication cost.}
Compared with the fixed-bit algorithms, our proposed DQ-SGD can achieve the same performance with fewer communication costs. The communication cost of DQ-SGD and fixed-bit algorithms are shown as follows with proofs given in Appendix \ref{pro:Theorem 2}.

\begin{theorem}
For the problem in Eq. \eqref{optim_problem} under Assumptions \ref{ass:stochastic_gradient}, \ref{ass:smoothnesee}, \ref{ass:strongly convexity}, with initial parameter $\mathbf{x}_0$, using the dynamic quantizer in Eqs. \eqref{eq:bits} to quantize the gradients, then the total communication cost for DQ-SGD is upper bounded by
\begin{equation}
\begin{split}
    f_{\mathcal{Q}}(T, W, b_t) \le  WdT\log_2{\sqrt{\cfrac{T\left[2L[F(\mathbf{x}_0)-\mathbf{x}^*] + \sigma^2\right]}{{\hat \epsilon}_Q}}}\\
    + WTB_{pre} + WTd + \cfrac{WTd}{2}\log_2{GM(\alpha)}\\
\end{split}
\end{equation}

If we want to achieve the same model performance, the total communication cost of the fixed-bit algorithms is upper bounded by
\begin{equation}
\begin{split}
    f_{\mathcal{Q}}(T, W, b_t) \le  WdT\log_2{\sqrt{\cfrac{T\left[2L[F(\mathbf{x}_0)-\mathbf{x}^*] + \sigma^2\right]}{{\hat \epsilon}_Q}}}\\
    + WTB_{pre} + WTd + \cfrac{WTd}{2}\log_2{AM(\alpha)}\\
\end{split}
\end{equation}
where Arithmetic Mean $AM(\alpha) = \cfrac{1}{T}\sum_{t=0}^{T-1} \alpha^{t} = \cfrac{1}{T} \cfrac{1-\alpha^{T}}{1-\alpha}$ and Geometric Mean $GM(\alpha) = \left[\prod_{t=0}^{T-1} \alpha^{t}\right]^{\frac{1}{T}} = \alpha^{\frac{T-1}{2}}$.
\label{Theorem 2}
\end{theorem}

We can see that if we desire a lower quantization error, we need more communication costs. Note that $0 <\alpha <1$, so $AM(\alpha) > GM(\alpha)$, which means our proposed DQ-SGD uses fewer communication cost compared with the fixed-bit algorithms. 

\begin{algorithm}[!htp]
\label{alg:DQSGD in distributed learning}
\caption{DQ-SGD in Distributed Learning}
\LinesNumbered 
\KwIn{Iterations number $T$, desired quantization error upper bound $\epsilon_Q$, learning rate $\eta$, initial point $\mathbf{x}_0 \in \mathbb{R}^d$, initial number of quantization bits $b_0$, hyper-parameters $\alpha$}
\KwOut{$\mathbf{x}_T$}
\For{$t=0, 1,..., T-1$}{

	\textbf{On each worker {$l=1, ..., W$}:}\\

	Compute local gradient $\mathbf{g}^{(i)}_t$\;

	Quantize the gradient $\mathcal {Q}_{b_t}[\mathbf{g}^{(i)}_t]$ according to Eq.~\eqref{quantizedoperation}\;

	Send $\mathcal {Q}_{b_t}[\mathbf{g}^{(i)}_t]$ to server\;

	Receive $\mathbf{x}_{t+1}$ and $b_{t+1}$ from server\;

	\textbf{On server:}\\
	Collect all $W$ quantized gradients $\mathcal {Q}_{b_t}[\mathbf{g}^{(i)}_t]$ from workers\;

	Average: $\mathbf{\hat g}_t = \frac{1}{W}\sum_{l=1}^W \mathcal {Q}_{b_t}[\mathbf{g}^{(i)}_t]$\;
	
	Update the global parameters $\mathbf{x}_{t+1} = \mathbf{x}_t - \eta \mathbf{\hat g}_t$\;
	
    Update the quantization bits $b_{t+1}$ according to Eq.~\eqref{eq:bits}\;

	Send $\mathbf{x}_{t+1}$ and $b_{t+1}$ to all workers\;}
\end{algorithm}

%% file: 6.5.discussion.tex
\section{Discussions}
\label{sec:discussion}

\subsection{Convergence Error for Quadratic Objectives.}
In previous sections, we use the upper bound of convergence error to measure the model performance. In this subsection, we will prove that there exist strongly convex functions $F(\mathbf{x})$ where the convergence error bound in Theorem \ref{Theorem 1} is tight (i.e., The `$=$' in Eq. \eqref{eq:Theorem 1} can be achieved.). 

For general quadratic functions, we can employ gradient flow\footnote{when the 
learning rate is infinitesimal, the stochastic gradient descent process can be regarded as a stochastic dynamic system.} to calculate an exact convergence error. We have the relationship between the aggregated stochastic gradients and full gradients: ${\mathbf{\hat g}}_t=\nabla F(\mathbf{x}_t)+\bm{\epsilon}_t$. Based on the central limit theorem, it is assumed that $\bm{\epsilon}_t$ follows the Gaussian distribution, that is $\bm{\epsilon}_t\sim \mathcal{N}(\bm{0},\bm{\Sigma}(\mathbf{x}_t))$. Then using analysis within the gradient flow framework, we can get the following theorem.

\begin{theorem}[Exact Convergence Error for Quadratic Objectives]
	For a quadratic optimization objective function $F(\mathbf{x}) =  1/2 \mathbf{x}^\mathrm{T}\mathbf{H} \mathbf{x} + \mathbf{A}^\mathrm{T} \mathbf{x} + B$, consider the perturbed gradient descent dynamics
\begin{equation}
\mathbf{x}_{t+1} = \mathbf{x}_t - \eta \nabla F(\mathbf{x}_t) - \eta \bm{\epsilon}_t, \bm{\epsilon}_t \sim \mathcal{N}(\bm{0},\bm{\Sigma}(\mathbf{x}_t))
\end{equation}
We can achieve 
\begin{align}
	&\mathbb{E}[F(\mathbf{x}_T)-F(\mathbf{x}^*)]\nonumber\\
	&=\cfrac{1}{2} (\mathbf{x}_0-\mathbf{x}^*)^\text{\rm T} (\bm{\rho}(\eta)^T)^\text{\rm T} \mathbf{H} \bm{\rho}(\eta)^T (\mathbf{x}_0-\mathbf{x}^*)\nonumber\\
	 &+\cfrac{\eta^2}{2}\sum_{t=0}^{T-1}  \text{\rm Tr} \left[ \bm{\rho}(\eta)^{T-1-t} \bm{\Sigma}(\mathbf{x}_t) \mathbf{H} \left(\bm{\rho}(\eta)^{T-1-t}\right)^\text{\rm T}\right]
	 \label{eq:quandratic}
\end{align}
where $\bm{\rho}(\eta) : = \mathbf{I}-\eta \mathbf{H}$, and $\mathbf{H}$ is the Hessian matrix.
\label{Theorem 3}
\end{theorem}
Detailed proof is in Appendix \ref{pro:Theorem 3}. We can see that the convergence error consists of two parts: the error of the gradient descent method, which is linearly convergent; the error due to gradient estimation error (data sampling noise, gradient quantization error).

Consider the case where the Hessian matrix is isotropic $\mathbf{H} = \lambda \mathbf{I}$, and let $\beta(\eta) := 1 - 2 \eta \lambda + \eta^2 \lambda^2$, then Eq.\eqref{eq:quandratic} can be rewrite as
\begin{align}
\mathbb{E}[F(\mathbf{x}_T)-F(\mathbf{x}^*)] = \beta(\eta)^T [F(\mathbf{x}_0)-F(\mathbf{x}^*)] \nonumber\\
+ \cfrac{\lambda \eta^2}{2} \sum_{t=0}^{T-1} \beta(\eta)^{T-1-t} \text{Tr}[\bm{\Sigma}(\mathbf{x}_t)]
\label{eq:quaDiff}
\end{align}

In Lemma \ref{lem:qsg}, if we let the gradient noise always reach the upper bound value, then we can get
\begin{align}
  \text{Tr}[\bm{\Sigma}(\mathbf{x}_t)]&= \mathbb{E}\left[\|\mathbf{\hat g}_t - \nabla F(\mathbf{x}_t)\|_2^2 \right]\notag\\
  &= \cfrac{\sigma^2}{W} +\cfrac{d}{4W(2^{b_t-1}-1)^2}\bar G_t  
\label{eq:cov}
\end{align}
Plugging Eq.~\eqref{eq:cov} into Eq.~\eqref{eq:quaDiff}, then the `$=$' in Eq. \eqref{eq:Theorem 1} can be achieved, and proves that the upper bound for strongly convex objectives in Theorem \ref{Theorem 1} is tight in this case. 

\subsection{Algorithm Implementation Details}
Although Eq. \eqref{eq:bits} provide valuable insights about how to adjust $b_t$ over time, it is still challenging to use it in practice due to the convergence rate $\alpha$ being known. Inspired by \cite{b19}, we propose a straightforward rule where we approximate $F(\mathbf{x}^*)$ to 0 and the learning rate $\eta$ is small enough. We estimate $\alpha$ as follows according to Theorem \ref{Theorem 1}:
\begin{equation}
	\alpha_{est} = \left[\cfrac{F(\mathbf{x}_t)}{F(\mathbf{x}_0)}\right]^{1/t}
\end{equation}

where $F(\mathbf{x}_0)$ and $F(\mathbf{x}_t)$ can be easily obtained in the training.

\subsection{Dynamic Adjustment in the Number of Bits}
From Eq.~\eqref{eq:bits}, we can see that two factors may affect how to adjust the number of quantization bits: the increased weight $\alpha^{(T-1-t)/2}$ and the root-mean-square of local gradient norm $\bar G_t$.
$\alpha^{(T-1-t)/2}$ increases with the iterations $t$, and $\bar G_t$ gets smaller as the training process goes on. Therefore, 

$\bullet$ {\bf Decreasing in Communication.} If the decreasing rate of the root-mean-square of local gradient norm (i.e., $\frac{\bar G_{t+1}}{\bar G_t}$) smaller than $\sqrt{\alpha}$, then $b_{t+1}<b_t$, which means the number of quantization bits decreases with the iteration step; 

$\bullet$ {\bf Increasing in Communication.} On the contrary, if the decreasing rate $\frac{\bar G_{t+1}}{\bar G_t}$ is bigger than $\sqrt{\alpha}$, then $b_{t+1}>b_t$, meaning that the number of quantization bits increases with the iteration step.

%% file: 7.experiments.tex
\section{Experiments}
\label{sec:experiments}
\begin{figure*}[!htbp]
	\centering
	\subfigure[AG-News (SGD:1313.29GB, 0.9016)]{
		\includegraphics[width=0.60\columnwidth]{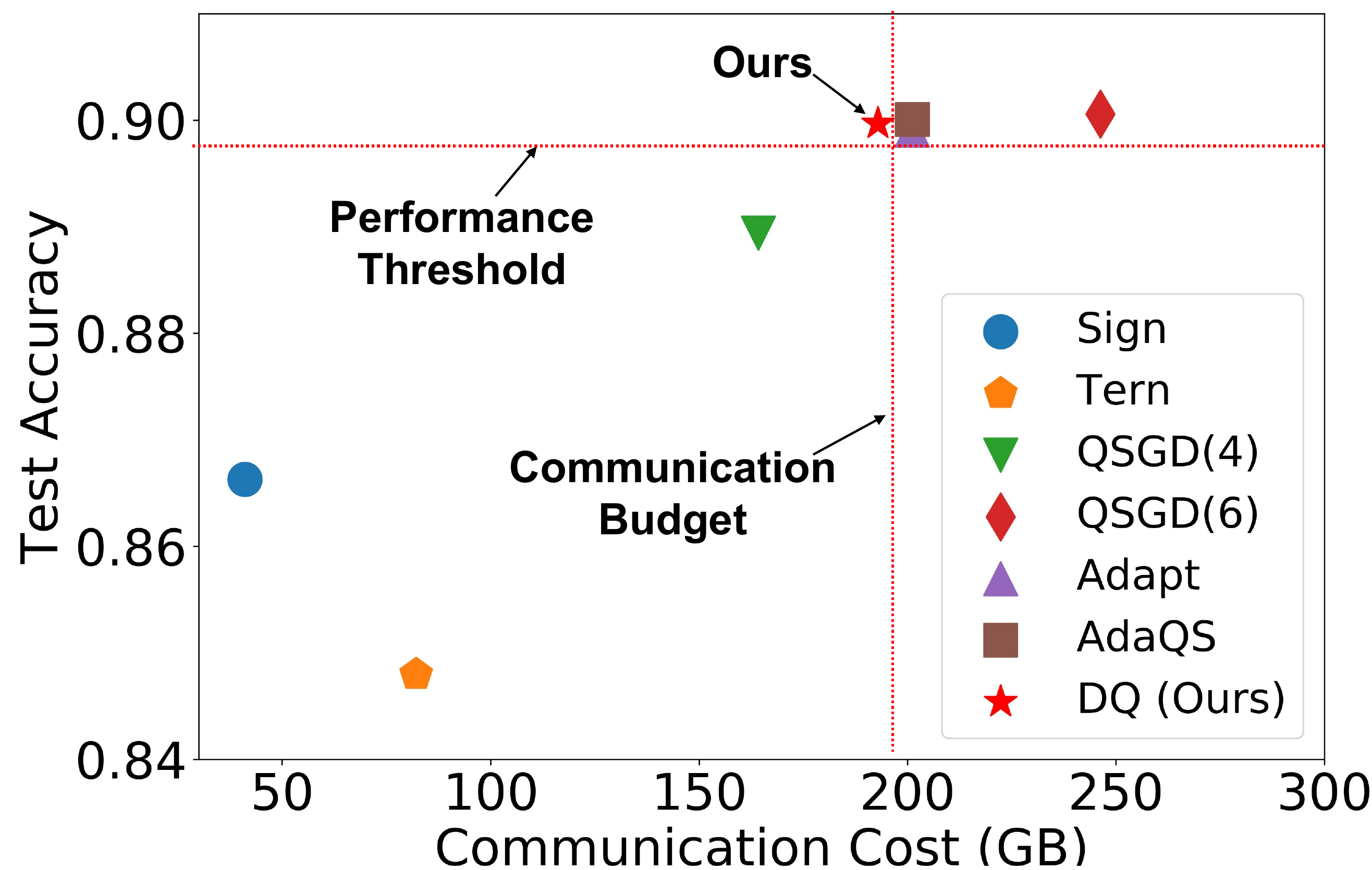}
	}
	\quad
	\subfigure[CIFAR10 (SGD:1998.06GB, 0.8815)]{
		\includegraphics[width=0.58\columnwidth]{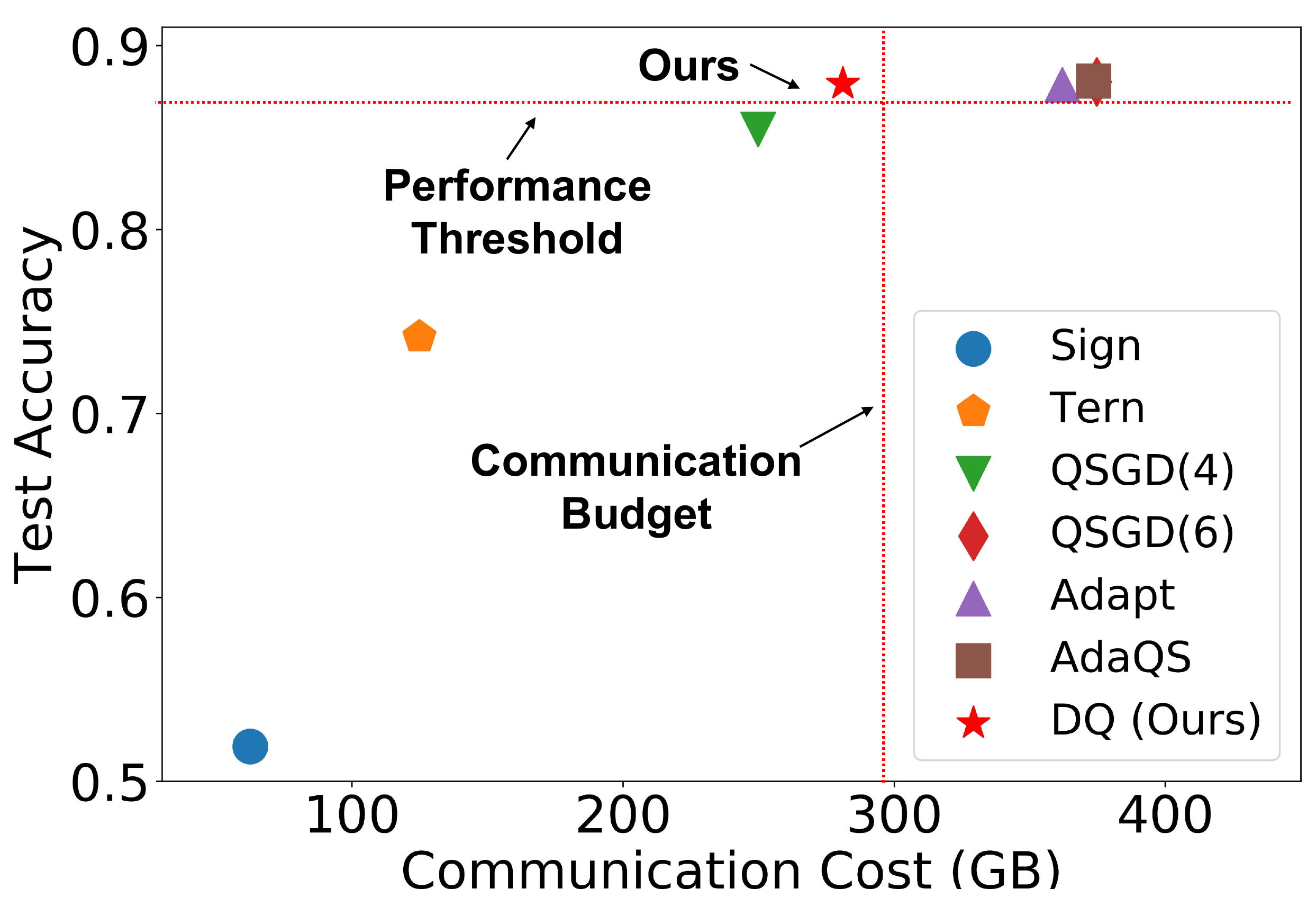}
	}
	\quad
	\subfigure[CIFAR100 (SGD:3805.54GB, 0.6969)]{
		\includegraphics[width=0.60\columnwidth]{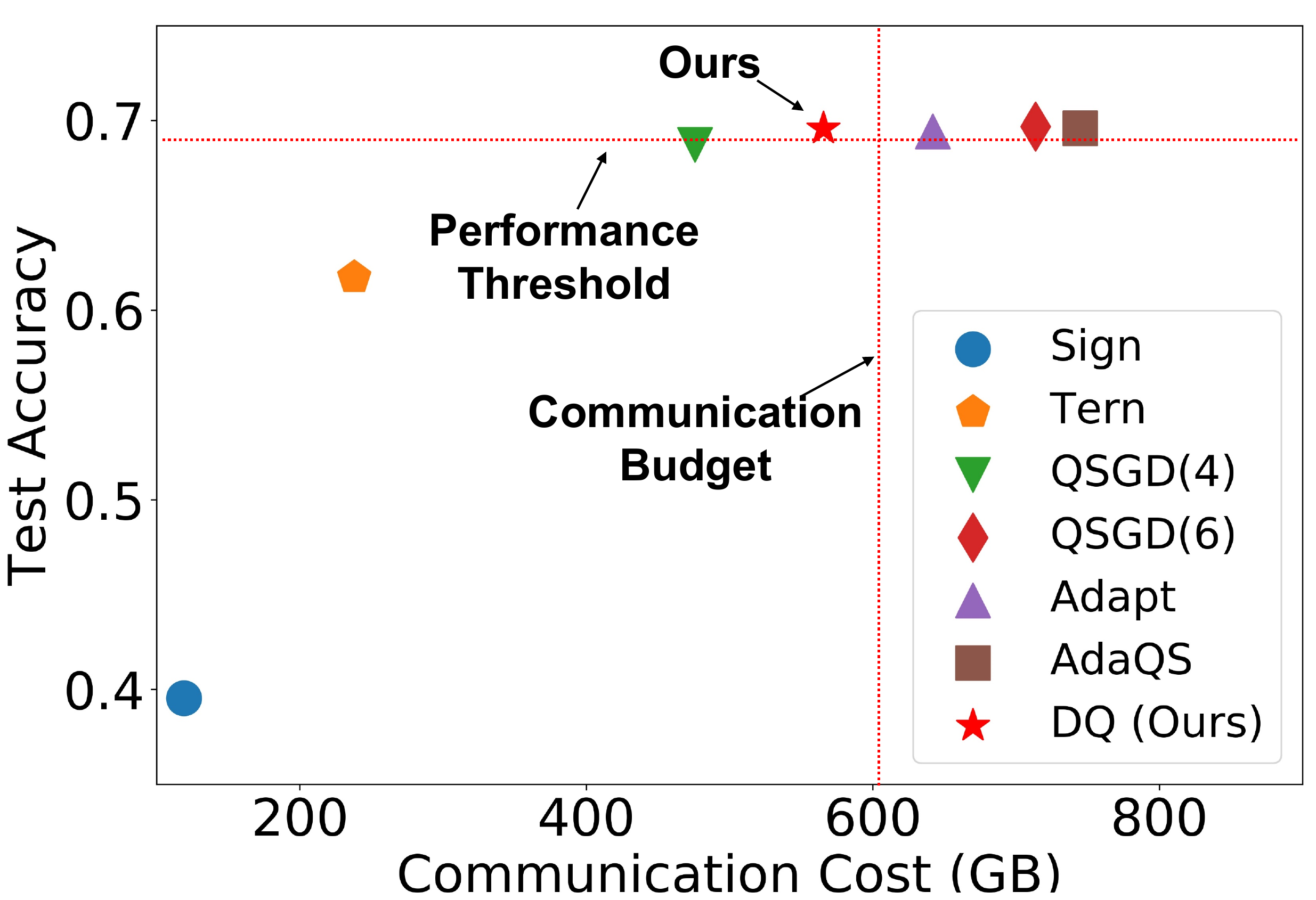}
	}

	\caption{Performance comparison with SOTA on AG-News, CIFAR-10, and CIFAR-100.}
	\label{fig:sota}
	\vspace{-0.2in}
\end{figure*}

\begin{figure*}[!htbp]
	\centering
	\subfigure[Test accuracy]{
		\includegraphics[width=0.60\columnwidth]{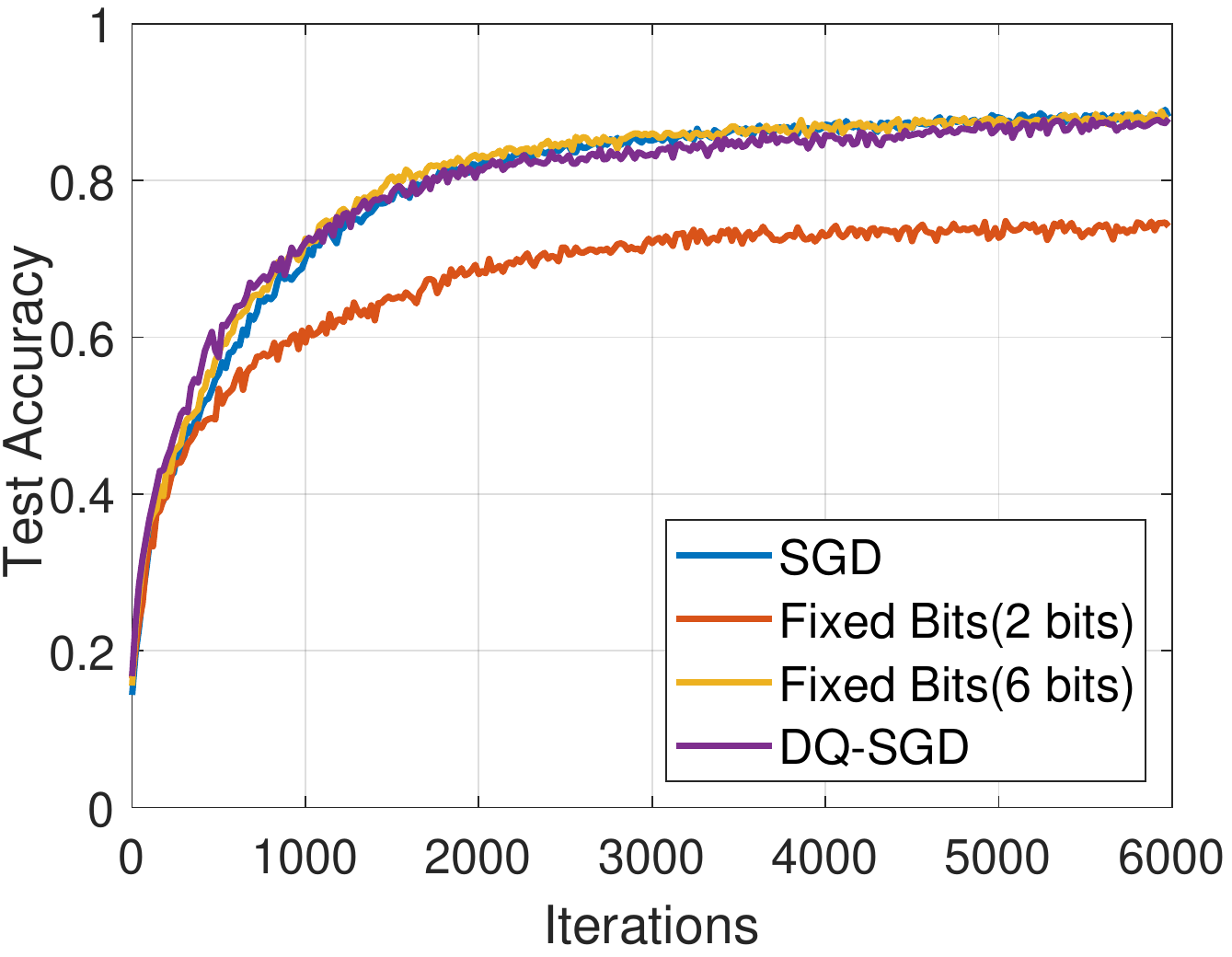}
	}
	\quad
	\subfigure[Training loss]{
		\includegraphics[width=0.6\columnwidth]{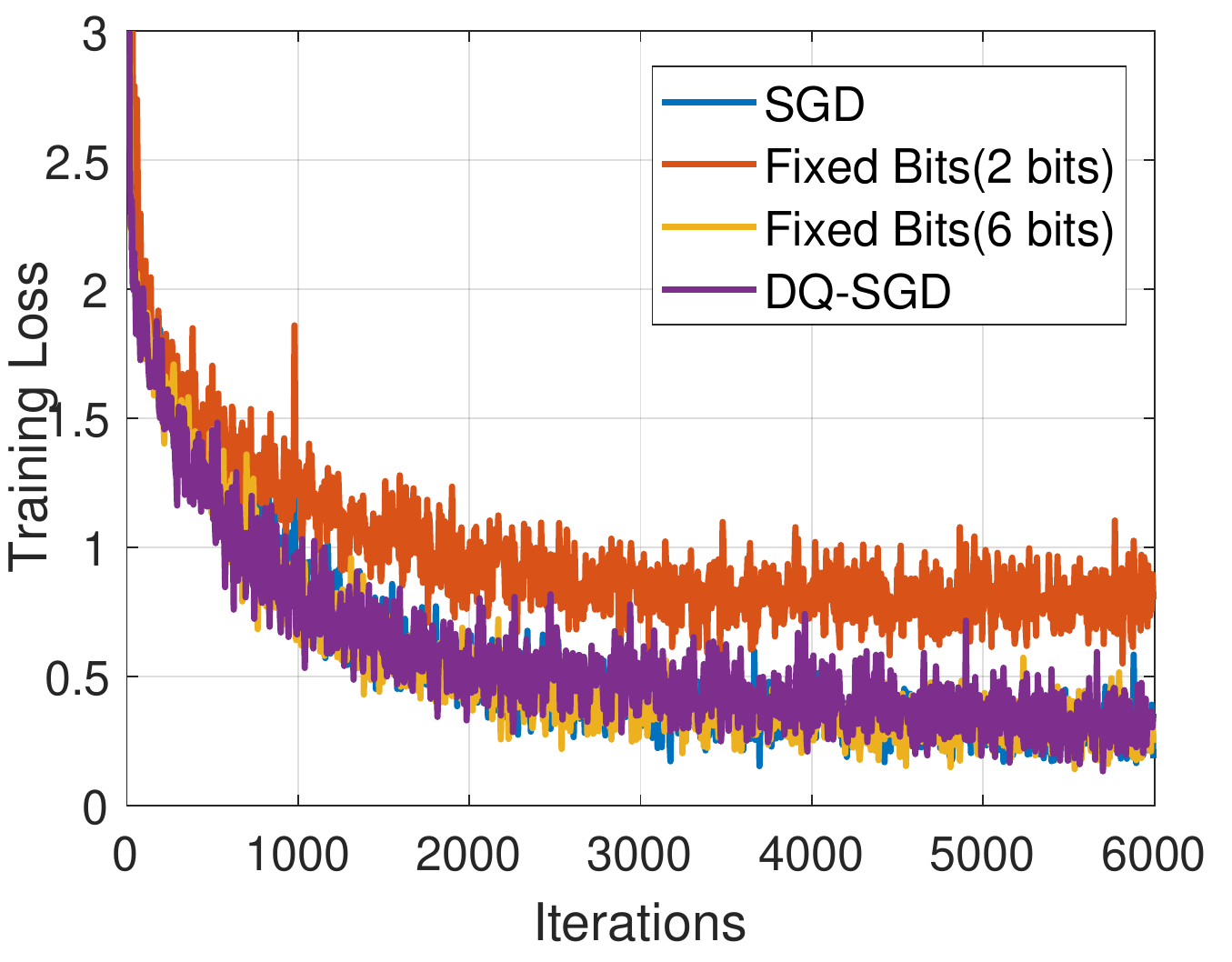}
	}
	\quad
	\subfigure[Bits allocation]{
		\includegraphics[width=0.6\columnwidth]{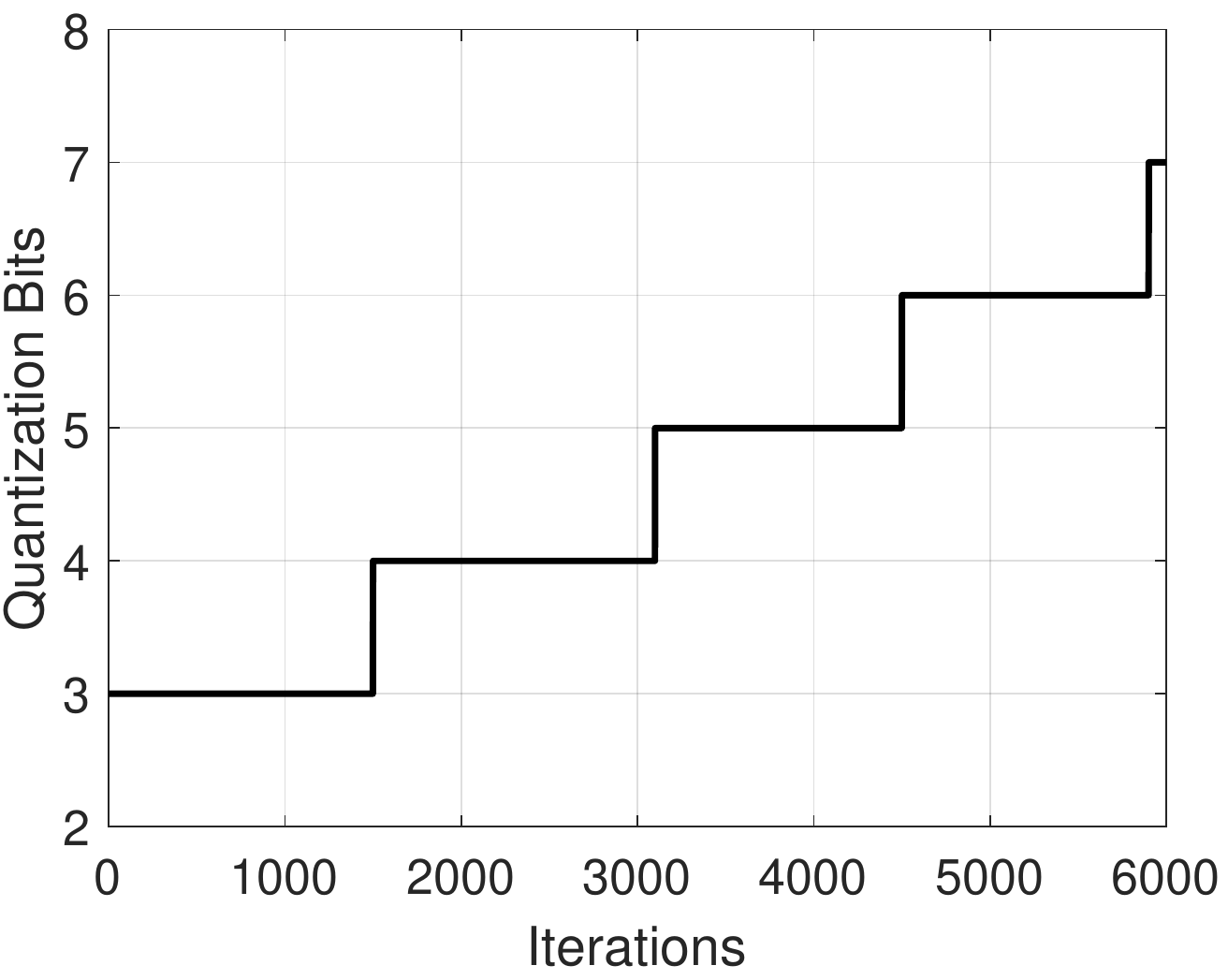}
	}
	\caption{The learning process of Fixed Bits and DQ-SGD on CIFAR-10.}
	\label{fig:comparison_results}
	\vspace{-0.1in}
\end{figure*}

\begin{figure}[!htbp]
	\centering
	\includegraphics[width=0.96\columnwidth]{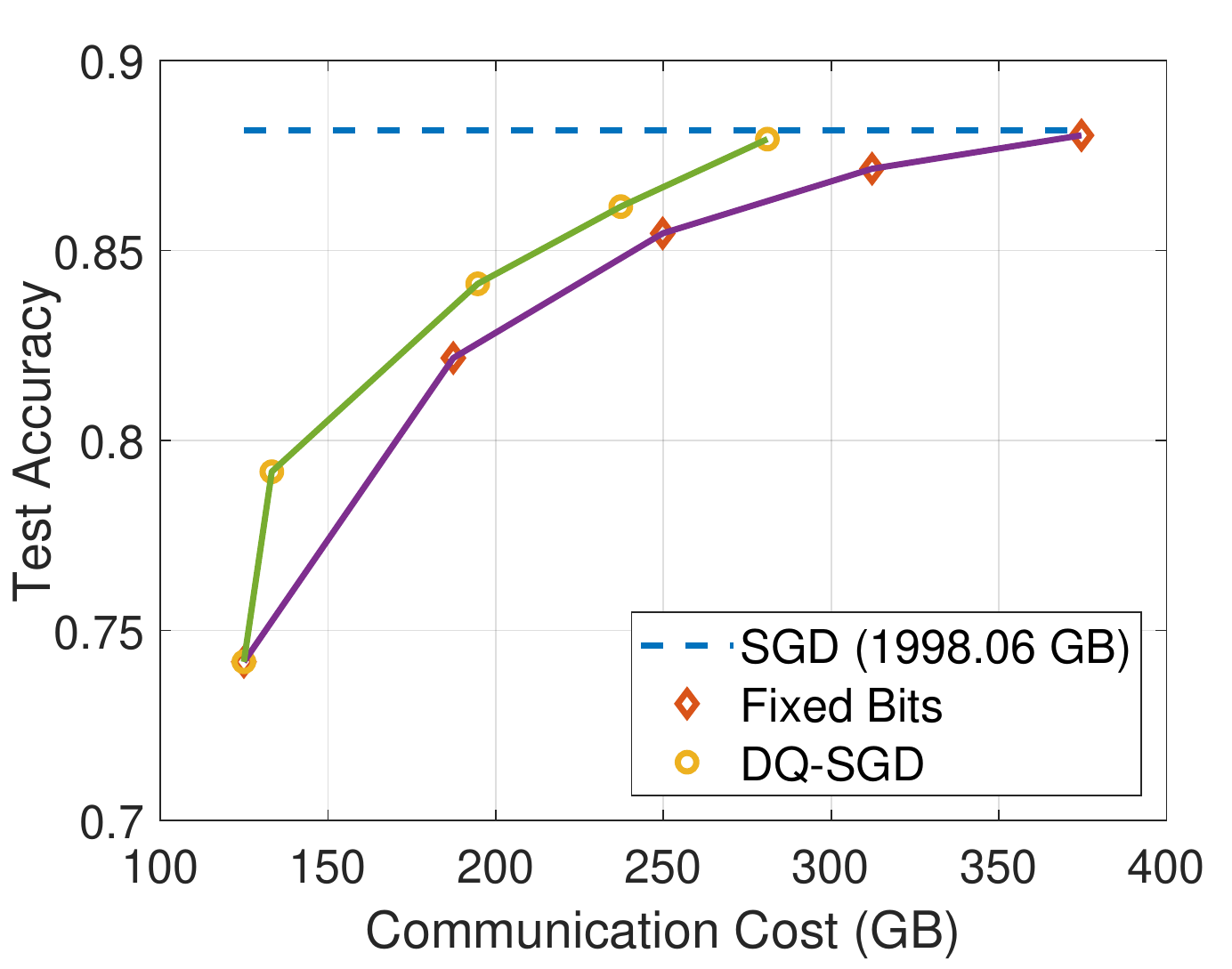}
	\caption{Test accuracy of Fixed Bits and DQ-SGD under different communication cost on CIFAR-10.}
	\label{fig:acc-ratio}
	\vspace{-0.2in}
\end{figure}

In this section, we conduct experiments on NLP and CV tasks on three datasets: AG-News \cite{b15}, CIFAR-10 \cite{b16}, and CIFAR-100 \cite{b16}, to validate the effectiveness of our proposed DQ-SGD method. We conduct experiments for $W = 8$ workers and use canonical networks to evaluate the performance of different algorithms: BiLSTM on the text classification task on the AG-News dataset, Resnet18 on the image classification task on the CIFAR-10 dataset, and Resnet34 on the image classification task on the CIFAR-100 dataset. Other parameters information is shown in Table \ref{tab:parameter}. We use test accuracy to measure the learning performance. We compare our proposed DQ-SGD with the following baselines: SignSGD \cite{b7}, TernGrad \cite{b14}, QSGD \cite{b4}, Adaptive \cite{b13} and AdaQS \cite{b11}.
\begin{table*}[!htbp]
	\caption{Parameters}
	\label{tab:parameter}
	\centering
	\begin{tabular}{|c|c|c|c|c|c|}
		\hline
		Dataset & Net & Learning rate & Batchsize & Interations & Hyperparameters in DQSGD\\ 
		\hline
		AG-News & BiLSTM & 0.005 & 32 & 1000 & $k=5, \alpha = 0.994$\\
		\hline
		CIFAR-10 & Resnet18 & 0.1 & 32 & 6000 & $k=20, \alpha = 0.999$\\
		\hline
		CIFAR-100 & Resnet34 & 0.01 & 64 & 6000 & $k=10, \alpha = 0.999$\\
		\hline
	\end{tabular}
\end{table*} 

\textbf{Test Accuracy vs Communication Cost}. Figure \ref{fig:sota} and table \ref{tab:acc-comratio} compare the test accuracy and communication cost of different algorithms under different tasks. The communication cost of Vanilla SGD are 1313.29 GB, 1998.06 GB, and 3805.54 GB, and the test accuracy can be achieved are 0.9016, 0.8815, and 0.6969 on AG-News, CIFAR-10, CIFAR-100, respectively. We set a communication budget as $15\%$ of the communication cost incurred by the SGD and a performance threshold as $99.7\%$ of the test accuracy achieved by the SGD for all three tasks. We can see that our proposed algorithm is the only one that satisfies a high-performance and low communication cost (the upper left region). Other baselines cannot achieve the performance threshold given the communication budget.

\textbf{Fixed Quantization vs. Dynamic Quantization}. Figure \ref{fig:comparison_results} shows the comparison results of the Fixed Bits algorithm and our proposed DQ-SGD on CIFAR-10. Figure 2 (a) and Figure 2 (b) show the test accuracy curves and the training loss curves. Figure 2 (c) shows the bits allocation of each iteration of DQ-SGD. Fixed Bits (6 bits) and DQ-SGD can get almost the same accuracy as SGD. However, the communication cost of DQ-SGD is reduced up to $25\%$ compared with that of Fixed Bits (6 bits). It can be seen that our dynamic quantization strategy can effectively reduce the communication cost compared with the fixed quantization scheme.  Figure~\ref{fig:acc-ratio} shows the accuracy of Fixed Bits and DQ-SGD under different communication costs. It can be seen that DQ-SGD can achieve higher test accuracy than Fixed Bits under the same communication cost. 

\begin{table*}[!ht]
 	\caption{Test accuracy vs. communication cost. 
	}
	\label{tab:acc-comratio}
	\centering
			\begin{tabular}{|c|c|c|c|c|c|c|}
				\hline
				{} &\multicolumn{2}{|c|}{AG-News} &\multicolumn{2}{|c|}{CIFAR-10} &\multicolumn{2}{|c|}{CIFAR-100}\\
				\hline
				{} & \tabincell{c}{Accuracy} & \tabincell{c}{Communication\\ Cost} & \tabincell{c}{Accuracy} & \tabincell{c}{ Communication\\ Cost} & \tabincell{c}{Accuracy} & \tabincell{c}{Communication\\ Cost}\\
				\hline
				Vanilla SGD &0.9016 &1313.29 &0.8815 &1998.06 &0.6969 &3805.54\\
				\hline
				SignSGD &0.8663 &41.04 &0.5191 &62.44 &0.3955 &118.92\\
				\hline
				TernGrad &0.8480 &82.08 &0.7418 &124.88 &0.6174 &237.85\\
				\hline
				QSGD (4 bits) &0.8894 &164.16 &0.8545 &249.76 &0.6837 &475.69\\
				\hline
				\hline
				QSGD (6 bits) &0.9006 &246.24 &0.8803 &374.64 &0.6969 &713.54\\
				\hline
				Adaptive &0.8991 &201.12 &0.8787 &361.97 &0.6943 &641.74\\
				\hline
				AdaQS &0.9001 & 201.13 &0.8809 &373.47 &0.6960 &744.72\\
				\hline
				DQSGD (Ours) &0.8997 &$\bm{192.85}$ &0.8793 &$\bm{280.98}$ &0.6959 &$\bm{565.46}$\\
				\hline
		\end{tabular}
\end{table*}

%% file: 8.conclusion.tex
\section{Conclusion}
\label{sec:conclusion}

This paper proposes a novel adaptive gradient quantization strategy called DQ-SGD to reduce the communication cost of distributed computing based on theoretical analysis. DQ-SGD adjusts quantization bits automatically by considering the norm of gradient and current iteration number. The experimental results of image classification and text classification show that DQ-SGD is superior to state-of-the-art gradient quantization methods in reducing communication costs.

%% file: Acknowledgment.tex
\section{Acknowledgment}
Prof. Linqi Song was supported in part by the Hong Kong RGC grant ECS 21212419, the Guangdong Basic and Applied Basic Research Foundation under Key Project 2019B1515120032, and the City University of Hong Kong SRG-Fd Grant 7005561.

%% file: A.appendix.tex
\appendices
\section{Proof of Lemma 1}
\label{pro: lem1}

According to Eq.~\eqref{eq:map}, we have
\begin{align*}
	\mathbb{E}[\zeta(g_j,s)] &= \cfrac{l}{s} [1-s\cfrac{\vert g_j \vert}{\|\mathbf{g}\|_p}+l] +\cfrac{l+1}{s} [s\cfrac{\vert g_j \vert}{\|\mathbf{g}\|_p}-l]
	= \cfrac{\vert g_j \vert}{\|\mathbf{g}\|_p}
\end{align*}

Then, we have
\begin{align*}
	\mathbb{E}[\zeta(g_j,s)^2] &= \mathbb{E}[\zeta(g_j,s)]^2+\mathbb{V}[\zeta(g_j,s)]\\
	& = \cfrac{\vert g_j \vert^2}{\|\mathbf{g}\|_p^2}+\cfrac{1}{s^2}p_r(1-p_r)
	\le \cfrac{\vert g_j \vert^2}{\|\mathbf{g}\|_p^2}+\cfrac{1}{4s^2}
\end{align*}

Considering that $\mathcal{Q}_b(g_j) = \|\mathbf{g}\|_p \cdot \text{sgn}(g_j)\cdot \zeta(g_j,s)$, we have
\begin{align*}
	\mathbb{E}[\|\mathcal{Q}_b[\mathbf{g}]\|_2^2] &=\sum_{j=0}^{d} \mathbb{E}[\|\mathbf{g}\|_p^2\zeta(g_j,s)^2]\\
	&\le \sum_{j=0}^{d} \|\mathbf{g}\|_p^2(\cfrac{\vert g_j \vert^2}{\|\mathbf{g}\|_p^2}+\cfrac{1}{4s^2})
	= \|\mathbf{g}\|_2^2 +\cfrac{d}{4s^2}\|\mathbf{g}\|_p^2
\end{align*}

So we can get $\mathbb{E}[\mathcal{Q}_b[\mathbf{g}]] = \mathbf{g}$ and
\begin{equation}\nonumber
	\text{\rm Tr}\left\{ \mathbb{V}\left[ \mathcal{Q}_{b_t}[\mathbf{g}^{(i)}_t)]\right]\right\} = \mathbb{E}[\|\mathcal{Q}_b[\mathbf{g}]\|_2^2-\|\mathbf{g}\|_2^2]\le \cfrac{d \|\mathbf{g}\|_p^2}{4(2^{b-1}-1)^2}. 
\end{equation}

Hence, for the aggregated gradient ${\mathbf{\hat g}}_t \triangleq \cfrac{1}{W}\sum_{i=1}^W \mathcal {Q}_{b_t}[\mathbf{g}^{(i)}_t]$:
\begin{equation}\nonumber
	\mathbb{E}\left[{\mathbf{\hat g}}_t]\right] = \cfrac{1}{W}\sum_{i=1}^W \mathbb{E}[\mathbf{g}^{(i)}_t] \overset{(a)}{=} \nabla F(\mathbf{x}_t)
\end{equation}
\begin{align*}
    \mathbb{E}\left[\|{\mathbf{\hat g}}_t\|_2^2\right] &= \text{\rm Tr}\left\{\mathbb{V}\left[{\mathbf{\hat g}}_t\right]\right\}+ \mathbb{E}\left[{\mathbf{\hat g}}_t\right]^{\text{T}}\mathbb{E}\left[{\mathbf{\hat g}}_t\right]\\
  &=\cfrac{1}{W^2} \sum_{i=1}^W \text{\rm Tr}\left\{ \mathbb{V}\left[ \mathcal{Q}_{b_t}[\mathbf{g}^{(i)}_t)]\right]\right\} + \|\cfrac{1}{W}\sum_{i=1}^W \mathbf{g}^{(i)}_t\|_2^2\\
  & \le \cfrac{d}{4W^2(2^{b_t-1}-1)^2} \sum_{i=1}^W\|\mathbf{g}^{(i)}_t\|_p^2 + \|\cfrac{1}{W}\sum_{i=1}^W \mathbf{g}^{(i)}_t\|_2^2\\
  & \le \cfrac{d}{4W^2(2^{b_t-1}-1)^2} \sum_{i=1}^W\|\mathbf{g}^{(i)}_t\|_p^2 + \cfrac{1}{W^2}\sum_{i=1}^W \|\mathbf{g}^{(i)}_t\|_2^2 \\
  &\overset{(a)}{\le} \cfrac{d}{4W(2^{b_t-1}-1)^2}\bar G_t^2 + \|\nabla F(\mathbf{x}_t)\|_2^2 + \cfrac{\sigma^2}{W}
\end{align*}

where $(a)$ uses the Assumption \ref{ass:stochastic_gradient}, and $\bar G_t^2 = \cfrac{1}{W}\sum_{i=1}^W\|\mathbf{g}^{(i)}_t\|_p^2$.	

\section{Proof of Theorem 1}
\label{pro:Theorem 1}

Considering function $F$ is $L-\text{smooth}$, and using Assumption \ref{ass:smoothnesee}:
\begin{equation}\nonumber
\begin{split}
F(\mathbf{x}_{t+1}) \le F(\mathbf{x}_t) + \nabla F(\mathbf{x}_t)^\text{T} (\mathbf{x}_{t+1}-\mathbf{x}_t)
+ \cfrac{L}{2} \|\mathbf{x}_{t+1}-\mathbf{x}_t\|_2^2
\end{split}
\end{equation}

Due to $\mathbf{x}_{t+1} = \mathbf{x}_t -\cfrac{\eta}{W}\sum_{i=1}^W \mathcal{Q}_{b_t}[\mathbf{g}^{(i)}_t]$, then:
\begin{align*}
F(\mathbf{x}_{t+1})&\le F(\mathbf{x}_t) + \nabla F(\mathbf{x}_t)^\text{T} (-\cfrac{\eta}{W}\sum_{i=1}^W \mathcal{Q}_{b_t}[\mathbf{g}^{(i)}_t])\\ 
&\qquad + \cfrac{L}{2} \|-\cfrac{\eta}{W}\sum_{i=1}^W \mathcal{Q}_{b_t}[\mathbf{g}^{(i)}_t]\|_2^2
\end{align*}

Taking total expectations, and using Lemma \ref{lem:qsg}, this yields:
\begin{equation}\nonumber
\begin{split}
\mathbb{E}[F(\mathbf{x}_{t+1}) - F(\mathbf{x}_t)] \le (-\eta + \cfrac{L \eta^2}{2})\|\nabla F(\mathbf{x}_t)\|_2^2 \\  + \cfrac{L\eta^2\sigma^2}{2W}
+ \cfrac{L\eta^2d}{8W(2^{b_t-1}-1)^2}\bar G_t^2
\end{split}
\end{equation}

Considering that function $F$ is $\mu-\text{strongly convex}$, and using Assumption \ref{ass:strongly convexity}, so:
\begin{equation}\nonumber
\begin{split}
\mathbb{E}[F(\mathbf{x}_{t+1}) - F(\mathbf{x}_t)] \le  - (2\mu\eta-L\mu \eta^2)[F(\mathbf{x}_t)-F(\mathbf{x}^*)]\\  
+ \cfrac{L\eta^2\sigma^2}{2W}
+ \cfrac{L\eta^2d}{8W(2^{b_t-1}-1)^2}\bar G_t^2
\end{split}
\end{equation}

Subtracting $F(\mathbf{x}^*)$ from both sides, and let $\alpha(\eta) := 1-2\mu\eta+L\mu \eta^2$, so:
\begin{equation}\nonumber
\begin{split}
	\mathbb{E}[F(\mathbf{x}_{t+1})-F(\mathbf{x}^*)] \le \alpha[F(\mathbf{x}_t)-F(\mathbf{x}^*)]+ \cfrac{L\eta^2\sigma^2}{2W}\\   
    + \cfrac{L\eta^2d}{8W(2^{b_t-1}-1)^2}\bar G_t^2
\end{split}
\end{equation}

Applying this recursively, we conclude the proof.


\section{Proof of Theorem 2}
\label{pro:Theorem 2}

The bits allocation is:
$b_t \approx \log_2{\left[\sqrt{\cfrac{T}{{\hat \epsilon}_Q}}\alpha^{(T-1-t)/2}\bar G_t + 1\right]}+1$, so
\begin{equation}\nonumber
\begin{split}
    &f_{\mathcal{Q}}(T, W, b_t)= W\sum_{t=0}^{T-1} \left[db_t+B_{pre}\right]\\
    &\approx Wd\sum_{t=0}^{T-1}\log_2{\left[\sqrt{\cfrac{T}{{\hat \epsilon}_Q}}\alpha^{(T-1-t)/2}\bar G_t\right]}+WTB_{pre} + WTd\\
    &< WdT\log_2{\sqrt{\cfrac{T\bar G_0^2}{{\hat \epsilon}_Q}}} + \cfrac{WdT(T-1)}{4}\log_2{\alpha}+WTB_{pre} + WTd\\
    &< WdT\log_2{\sqrt{\cfrac{T\left[2L[F(\mathbf{x}_0)-\mathbf{x}^*] + \sigma^2\right]}{{\hat \epsilon}_Q}}} + WTB_{pre}\\
    &\qquad + WTd+\cfrac{WdT(T-1)}{4}\log_2{\alpha}\\
\end{split}
\end{equation}

Accordingly, if we fix the number of quantization bits (i.e., $b_t=b$), then we have 
\begin{equation}\nonumber
\begin{split}
    {\hat \epsilon}_Q &= \sum_{t=0}^{T-1} \alpha^{T-1-t} \cfrac{\bar G_t^2}{(2^{b_t-1}-1)^2}\\
    & \le \cfrac{\bar G_0^2}{(2^{b-1}-1)^2} \sum_{t=0}^{T-1} \alpha^{T-1-t} \\
    & \le \cfrac{2L[F(\mathbf{x}_0)-\mathbf{x}^*] + \sigma^2}{(2^{b-1}-1)^2}\cfrac{1-\alpha^{T}}{1-\alpha}\\
\end{split}
\end{equation}

So,
\begin{equation}\nonumber
\begin{split}
    b \le \log_2{\left[\sqrt{\cfrac{2L[F(\mathbf{x}_0)-\mathbf{x}^*] + \sigma^2}{{\hat \epsilon}_Q}\cfrac{1-\alpha^{T}}{1-\alpha}} + 1\right]}+1
\end{split}
\end{equation}

So, the total communication cost for the fixed bits algorithm is

\begin{equation}\nonumber
\begin{split}
    &f_{\mathcal{Q}}(T, W, b_t)= W\sum_{t=0}^{T-1} \left[db_t+B_{pre}\right]= WTdb + WTB_{pre}\\
    &= WTd\log_2{\left[\sqrt{\cfrac{2L[F(\mathbf{x}_0)-\mathbf{x}^*] + \sigma^2}{{\hat \epsilon}_Q}\cfrac{1-\alpha^{T}}{1-\alpha}} + 1\right]}\\
    &\qquad + WTB_{pre}+WTd\\
    &\approx WTd\log_2{\sqrt{\cfrac{2L[F(\mathbf{x}_0)-\mathbf{x}^*] + \sigma^2}{{\hat \epsilon}_Q}}} + WTB_{pre}\\
    &\qquad +WTd + WTd\log_2{\sqrt{\cfrac{1-\alpha^{T}}{1-\alpha}}}\\
\end{split}
\end{equation}

If we let $AM(\alpha) = \cfrac{1}{T}\sum_{t=0}^{T-1} \alpha^{t} = \cfrac{1}{T} \cfrac{1-\alpha^{T}}{1-\alpha}$ and $GM(\alpha) = (\prod_{t=0}^{T-1} \alpha^{t})^{(1/T)} = \alpha^{(T-1)/2}$.

Then the total communication cost for DQ-SGD is
\begin{equation}\nonumber
\begin{split}
    f_{\mathcal{Q}}(T, W, \{b_t\}) \le  WdT\log_2{\sqrt{\cfrac{T\left[2L[F(\mathbf{x}_0)-\mathbf{x}^*] + \sigma^2\right]}{{\hat \epsilon}_Q}}}\\
    + WTB_{pre} + WTd + \cfrac{WTd}{2}\log_2{GM(\alpha)}\\
\end{split}
\end{equation}

and the total communication cost for fixed bits is
\begin{equation}\nonumber
\begin{split}
    f_{\mathcal{Q}}(T, W, \{b_t\}) \le  WdT\log_2{\sqrt{\cfrac{T\left[2L[F(\mathbf{x}_0)-\mathbf{x}^*] + \sigma^2\right]}{{\hat \epsilon}_Q}}}\\
    + WTB_{pre} + WTd + \cfrac{WTd}{2}\log_2{AM(\alpha)}\\
\end{split}
\end{equation}

\section{Proof of Theorem 3}
\label{pro:Theorem 3}

For a quadratic optimization problem $F(\mathbf{x}) =  1/2 \mathbf{x}^\mathrm{T}\mathbf{H} \mathbf{x} + \mathbf{A}^\mathrm{T} \mathbf{x} + B$, we consider a Gaussian noise case
\begin{equation}\nonumber
\mathbf{x}_{t+1} = \mathbf{x}_t - \eta \nabla F(\mathbf{x}_t) - \eta \bm{\epsilon}_t, \bm{\epsilon}_t \sim \mathcal{N}(\bm{0},\bm{\Sigma}(\mathbf{x}_t))
\end{equation}

Then we have
\begin{equation}\nonumber
\begin{split}
	\mathbf{x}_{t+1}&=\mathbf{x}_t-\eta \nabla F(\mathbf{x}_t)-\eta \bm{\epsilon}_t\\
	&=\mathbf{x}_t-\eta [\mathbf{H} \mathbf{x}_t+\mathbf{A}]-\eta \bm{\epsilon}_t\\
	&=(\mathbf{I}-\eta \mathbf{H})\mathbf{x}_t-\eta \mathbf{A}-\eta \bm{\epsilon}_t
\end{split}
\end{equation}

Considering $\nabla F(\mathbf{x}^*) = \eta \mathbf{A} + \eta \mathbf{H} \mathbf{x}^* = 0$, subtracting $\mathbf{x}^*$ from both sides, and rearranging, this yields:
\begin{equation}\nonumber
\begin{split}
	\mathbf{x}_{t+1}-\mathbf{x}^*&=(\mathbf{I}-\eta \mathbf{H})\mathbf{x}_t-\eta \mathbf{A}-\mathbf{x}^*-\eta \bm{\epsilon}_t\\
	&=(\mathbf{I}-\eta \mathbf{H})(\mathbf{x}_t-\mathbf{x}^*)-\eta \mathbf{A}-\eta \mathbf{H} \mathbf{x}^*-\eta \bm{\epsilon}_t\\
	&=(\mathbf{I}-\eta \mathbf{H})(\mathbf{x}_t-\mathbf{x}^*)-\eta \bm{\epsilon}_t
\end{split}
\end{equation}

Applying this recursively, let $\bm{\rho}=\mathbf{I}-\eta \mathbf{H}$, we have:
\begin{equation}\nonumber
	\mathbf{x}_T-\mathbf{x}^*=\bm{\rho}^T (\mathbf{x}_0-\mathbf{x}^*)- \sum_{t=0}^{T-1} [\eta \bm{\rho}^{T-1-t} \bm{\epsilon}_t]
\end{equation}

Considering that $\bm{\epsilon}_t \sim \mathcal{N}(\bm{0},\bm{\Sigma}(\mathbf{x}_t))$, then:
\begin{align*}
	&\sum_{t=0}^{T-1} [\eta \bm{\rho}^{T-1-t} \bm{\epsilon}_t] = \sum_{t=0}^{T-1} [\eta \bm{\rho}^{T-1-t} \bm{\Sigma}(\mathbf{x}_t)^{\frac 12} \mathcal{N}(\bm{0},\mathbf{I})]\\
	&= \sum_{t=0}^{T-1} [\eta \bm{\rho}^{T-1-t} \bm{\Sigma}(\mathbf{x}_t)^{\frac 12} [\mathbf{W}(t+1)-\mathbf{W}(t)]\}\equiv I(T)
\end{align*}
where, $\mathbf{W}$ is a standard $d\text{-dimensional}$ Wiener process, and $I(T)$ is an Ito integral. Hence $\mathbf{x}_T=\mathbf{x}^*+\bm{\rho}^T (\mathbf{x}_0-\mathbf{x}^*)-I(T)$, then:
\begin{equation}\nonumber
\begin{split}
	F(\mathbf{x}_T) &= \cfrac{1}{2}{\mathbf{x}_T}^\text{T}\mathbf{H} \mathbf{x}_T + \mathbf{A}^\text{T} \mathbf{x}_T + B\\
	&= \cfrac{1}{2}(\mathbf{x}^{(0)}-\mathbf{x}^*)^\text{T}(\bm{\rho}^T)^\text{T} \mathbf{H} \bm{\rho}^T (\mathbf{x}_0-\mathbf{x}^*) + \cfrac{1}{2} I(T)^\text{T} \mathbf{H} I(T) \\ &~~~~~+ F(\mathbf{x}^*)
	- [\bm{\rho}^T (\mathbf{x}_0-\mathbf{x}^*)+\mathbf{x}^*+\mathbf{A}]^\text{T} \mathbf{H} I(T)
\end{split}
\end{equation}

Subtracting $F(\mathbf{x}^*)$ from both sides, taking total expectations, and rearranging, this yields:
\begin{align*}
	&\mathbb{E}[F(\mathbf{x}_T)-F(\mathbf{x}^*)]\\
	&= \cfrac{1}{2} (\mathbf{x}_0-\mathbf{x}^*)^\text{T} (\bm{\rho}^T)^\text{T} \mathbf{H} \bm{\rho}^T (\mathbf{x}_0-\mathbf{x}^*)+ \cfrac{1}{2} \mathbb{E}[I(T)^\text{T} \mathbf{H} I(T)]\\
	& \quad -[\bm{\rho}^T (\mathbf{x}_0-\mathbf{x}^*)+\mathbf{x}^*+\mathbf{A}]^\text{T} \mathbf{H} \mathbb{E}[I(T)]
\end{align*}

The property of Ito integral $I(T)$ is:
\begin{equation}\nonumber
	\mathbb{E}[I(T)]=0
\end{equation}
\begin{align*}
	\mathbb{E}[I(T)^\text{T} \mathbf{H} I(T)]= \sum_{t=0}^{T-1} \eta^2 \text{Tr}[ \bm{\rho}^{T-1-t} \bm{\Sigma}(\mathbf{x}_t) \mathbf{H} (\bm{\rho}^{T-1-t})^\text{T}]
\end{align*}

Using this property, we have:
\begin{align*}
	\mathbb{E}[F(\mathbf{x}_T)-F(\mathbf{x}^*)]=\cfrac{1}{2} (\mathbf{x}_0-\mathbf{x}^*)^\text{T} (\bm{\rho}^T)^\text{T} \mathbf{H} \bm{\rho}^T (\mathbf{x}_0-\mathbf{x}^*)\\
	 +\cfrac{\eta^2}{2}\sum_{t=0}^{T-1}  \text{Tr} [ \bm{\rho}^{T-1-t} \bm{\Sigma}(\mathbf{x}_t) \mathbf{H} (\bm{\rho}^{T-1-t})^\text{T}]
\end{align*}

If we consider a simple example: the Hessian matrix is isotropic $\mathbf{H} = \lambda \mathbf{I}$, let $\alpha(\eta) := 1 - 2 \eta \lambda + \eta^2 \lambda^2$, so
\begin{align*}
	first &= \cfrac{1}{2} (\mathbf{x}_0-\mathbf{x}^*)^\text{T} (\bm{\rho}^T)^\text{T} \mathbf{H} \bm{\rho}^T (\mathbf{x}_0-\mathbf{x}^*)\\
	&= \alpha(\eta)^T \cfrac{1}{2} (\mathbf{x}_0-\mathbf{x}^*)^\text{T} \mathbf{H} (\mathbf{x}_0-\mathbf{x}^*)\\
	&=  \alpha(\eta)^T [\cfrac{1}{2} \mathbf{x}_0^\text{T} \mathbf{H} \mathbf{x}_0 + \cfrac{1}{2} \mathbf{x}^{*\text{T}} \mathbf{H} \mathbf{x}^* - \mathbf{x}_0^\text{T} \mathbf{H} \mathbf{x}^*]\\
	&= \alpha(\eta)^T [\cfrac{1}{2} \mathbf{x}_0^\text{T} \mathbf{H} \mathbf{x}_0 + \cfrac{1}{2} \mathbf{x}^{*\text{T}} \mathbf{H} \mathbf{x}^* - \mathbf{x}_0^\text{T} \mathbf{H} \mathbf{x}^* \\
	& \qquad + \mathbf{x}_0^\text{T} (\mathbf{H} \mathbf{x}^*+A) - \mathbf{x}^{*\text{T}} (\mathbf{H} \mathbf{x}^*+A)]\\
	&= \alpha(\eta)^T [\cfrac{1}{2} \mathbf{x}_0^\text{T} \mathbf{H} \mathbf{x}_0 - \cfrac{1}{2} \mathbf{x}^{*\text{T}} \mathbf{H} \mathbf{x}^* + \mathbf{x}_0^\text{T} A - \mathbf{x}^{*\text{T}} A]\\
	&= \alpha(\eta)^T [F(\mathbf{x}_0)-F(\mathbf{x}^*)]
\end{align*}

\begin{equation}\nonumber
second = \cfrac{\lambda \eta^2}{2} \sum_{t=0}^{T-1} \alpha(\eta)^{T-1-t} \text{Tr}[\bm{\Sigma}(\mathbf{x}_t)] 
\label{eq:qua_error1}
\end{equation}
Thus,
\begin{equation}\nonumber
\begin{split}
\mathbb{E}[F(\mathbf{x}_T)-F(\mathbf{x}^*)] = \alpha(\eta)^T [F(\mathbf{x}_0)-F(\mathbf{x}^*)]\\
+ \cfrac{\lambda \eta^2}{2} \sum_{t=0}^{T-1} \alpha(\eta)^{T-1-t} \text{Tr}[\bm{\Sigma}(\mathbf{x}_t)]
\end{split}
\end{equation}

\section{Proof of Eq.~(17)}
\label{pro:Corollary 1}

For the constrain function of Eq.~\eqref{eq:optimization}, we have
\begin{align*}
    &\cfrac{\partial^2[\sum_{t=0}^{T-1} \alpha^{T-1-t} \cfrac{\bar G_t^2}{(2^{b_t-1}-1)^2}]}{\partial b_t^2}\\
    &= \cfrac{(\ln{2})^2*(2^{b_t}+1)*2^{b_t}}{(2^{b_t-1}-1)^4}\alpha^{T-1-t}\bar G_t^2 > 0
\end{align*}
Therefore, this optimization problem is a convex optimization problem, then we have the Lagrange function
\begin{equation}\nonumber
\begin{split}
	\mathcal{L}(b_t, \lambda) = W\sum_{t=0}^{T-1} (db_t+B_{pre})\\
	+ \lambda \left[\sum_{t=0}^{T-1} \alpha^{T-1-t} \cfrac{\bar G_t^2}{(2^{b_t-1}-1)^2} - \epsilon_Q \right]
\end{split}
\end{equation}
where $\lambda$ is Lagrange multiplier. Then we can get
\begin{equation}\nonumber
	\cfrac{\partial \mathcal{L}(b_t, \lambda)}{\partial b_t} =Wd - \lambda \cfrac{2\ln2*2^{b_t-1}}{(2^{b_t-1}-1)^3} \alpha^{T-1-t}\bar G_t^2 = 0
\end{equation}
By solve this equation, we can get
\begin{equation}\nonumber
	b_t \approx \log_2{\left[\sqrt{\cfrac{2 \lambda \ln2}{Wd}}\alpha^{(T-1-t)/2}\bar G_t + 1\right]}+1
\end{equation}
Setting $\sum_{t=0}^{T-1} \alpha^{T-1-t} \cfrac{\bar G_t^2}{(2^{b_t-1}-1)^2}={\hat \epsilon}_Q$, we can solve $\lambda$ and get the result.